\documentclass[letterpaper, 10 pt, conference]{ieeeconf}

\IEEEoverridecommandlockouts

\title{\LARGE \bf
CompliantWBC: Whole-Body Compliance for Heavy Humanoids via Force Latent Estimation and Residual Impedance Targets
}

\author{Tan-Dzung Do$^{1*}$, Cuc T. Trinh$^{1*}$, Tuan Dat Phuong$^{2}$, Chien Le$^{1}$, Thanh Ly$^{1}$,\\
Vien Anh Ngo$^{1,3\dagger}$, and An Thai Le$^{1,3,4\dagger}$%
\thanks{$^{*}$Equal contribution. $^{\dagger}$Equal advising.}%
\thanks{$^{1}$VinRobotics, $^{2}$National University of Singapore, $^{3}$VinUniversity, $^{4}$TU Darmstadt.}%
}
\usepackage{mathtools}
\usepackage{amsmath}

\usepackage{amssymb}
\usepackage{amsfonts}
\usepackage{amsopn}
\usepackage{graphicx}
\usepackage{textcomp}
\usepackage{xfrac}
\usepackage{bbm}
\usepackage{overpic}
\usepackage{subfig}
\usepackage{wrapfig}
\usepackage{cuted}
\usepackage{capt-of}
\usepackage{algorithm}
\usepackage{algpseudocode}

\usepackage{multirow}

\usepackage{booktabs}
\usepackage{footnote}
\usepackage{paralist}
\let\labelindent\undefined
\usepackage{enumitem}
\usepackage{autobreak}
\usepackage{hyperref}
\usepackage{cleveref}
\crefname{appendix}{Appendix}{Appendices}
\Crefname{appendix}{Appendix}{Appendices}

\usepackage{cite}
\usepackage{siunitx}

\usepackage{color}
\definecolor{green}{rgb}{0, 0.4, 0}
\definecolor{orange}{rgb}{0.8, 0.6, 0.2}
\definecolor{red}{rgb}{1.0, 0.0, 0.0}
\definecolor{teal}{rgb}{0.0, 0.4, 0.4}
\definecolor{purple}{rgb}{0.65,0,0.65}
\definecolor{saffron}{rgb}{0.95,0.75,0.2}
\definecolor{turquoise}{rgb}{0.0,0.5,0.5}
\definecolor{brown}{rgb}{0.5, 0.16, 0.16}

\usepackage{overpic}
\usepackage{currfile}

\newlength\savedwidth

\definecolor{lightgray}{rgb}{0.6, 0.6, 0.6}

\newcommand{\addcite}[1]{{\textcolor{red}{[cite]}}}

\definecolor{revisedcolor}{RGB}{100,0,200}

\usepackage[normalem]{ulem}

\newcommand{\hidecomment}[1]{}
\newcommand{\R}{\mathbb{R}}

\usepackage{xspace}

\newcommand{\SE}{\mathrm{SE}}
\newcommand{\bvec}[1]{\boldsymbol{#1}}
\newcommand{\trans}{^{\top}}
\newcommand{\inv}{^{-1}}
\newcommand{\dsr}{_{d}}
\newcommand{\config}{\bvec{q}}
\newcommand{\vel}{\bvec{\nu}}
\newcommand{\mass}{\bvec{M}}
\newcommand{\torque}{\bvec{\tau}}
\newcommand{\jac}{\bvec{J}}
\newcommand{\force}{\bvec{f}}

\newcommand{\pose}{\bvec{x}}
\newcommand{\stiff}{\bvec{K}}
\newcommand{\damp}{\bvec{D}}

\newcommand{\latent}{\bvec{z}}
\newcommand{\prop}{\bvec{\mathrm{prop}}}
\newcommand{\tgt}[2]{\pose_{#1}(#2)}
\newcommand{\tgte}[1]{\tilde{\pose}_{#1}(\hat\force)}

\usepackage[table]{xcolor}

\begin{document}
\bstctlcite{IEEEexample:BSTcontrol}

\maketitle
\thispagestyle{empty}
\pagestyle{empty}

\begin{strip}
  \centering
  \includegraphics[width=\textwidth]{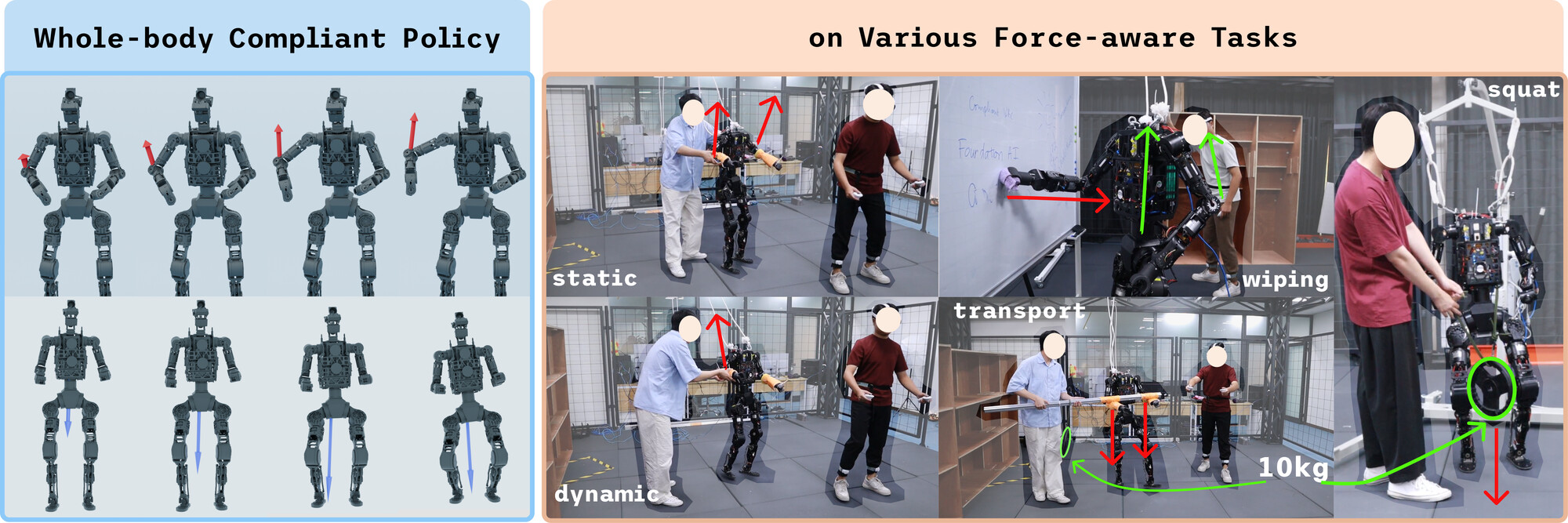}
  \captionof{figure}{\textsc{CompliantWBC} pairs a force-aware base policy with a
    bounded residual on the impedance equilibrium, trained under a Phong-weighted
    force-origin sampler that extends compliance to the whole body. We demonstrate
    it on a range of force-aware tasks on a real heavy humanoid.}
  \label{fig:teaser}
\end{strip}

\begin{abstract}

Whole-body compliant control is essential for deploying heavy humanoids under
high payload in human-centric environments. Most prior force-aware learning-based
pipelines focus on end-effector resistance, per-link upper-body springs, or
end-effector stiffness modulation, leaving \emph{arbitrary-site perturbations
on heavy platforms with lower-body engagement} largely unaddressed. We close this
gap with \textsc{CompliantWBC} comprising: (1)~A \emph{base policy} trained with RL to maximize compliance-fidelity reward, guided by a multi-site whole-body impedance \emph{reference controller}, extending classical Cartesian impedance to any controlled link; (2)~A bounded \emph{residual policy} that edits the per-link impedance
equilibrium over a frozen base, correcting the coarse but structured wrench estimate
supplied by a \emph{force encoder} co-trained behind a gradient barrier;
(3)~A \emph{Phong-weighted force-origin sampler} with an axis-decoupled pelvis
anchor induces lower-body-inclusive compliance curriculum training via two interpretable parameters. We evaluate \textsc{CompliantWBC} in
simulation against both compliant and stiff baselines, achieving best compliant fidelity of 2.58cm deviation from analytical solutions, and demonstrate it on a real
heavy humanoid across static/dynamic force reaction, board wiping, squat under
payload, and cooperative payload transport. Project website: \href{https://dotandung.github.io/compliantwbc/}{https://dotandung.github.io/compliantwbc/}

\end{abstract}
\section{Introduction}
\label{sec:intro}
Heavy humanoid loco-manipulation requires whole-body compliance. For instance, pelvis bracing against a wall, hip translation against a pulled cart, or double-support redistribution while lifting cannot be expressed by upper-body springs alone. Classical operational-space
methods~\cite{hogan1985impedance,ott2008cartesian,khatib1987unified,
sentis2005synthesis,henze2016ijrr,dietrich2016wbic,albuschaffer2007cartesian} cover this regime
analytically, but require accurate dynamics, hand-tuned task hierarchies,
and contact-mode reasoning that are brittle on heavy platforms.
Learning-based pipelines relax these requirements; however, previous works restrict compliance to the end-effector or to the upper-body
chain~\cite{zhang2025falcon,chen2025chip,xu2025facet,
lu2025gentlehumanoid,margolis2025softmimic}, leaving arbitrary-site multi-contact compliance with lower-body impedance participation
largely unaddressed.

In the learning-based context, the policy observes only proprioception and a motion command
(the external wrench $\force_{\text{ext}}$ is unobserved on hardware) and must
output torques that \emph{yield} compliantly to $\force_{\text{ext}}$ at
arbitrary contact sites while sustaining balance and command tracking.
Transferring classical compliance into a learning pipeline with reinforcement
learning (RL) therefore faces two main challenges. First, the classical impedance
formulation considers a single end-effector, so a principled
\emph{whole-body} heuristic must compose centroidal-momentum balance
with per-link impedance at arbitrary sites and distribute the resulting
wrench to a multi-contact support set. Second, training must actually exercise the
whole body (not just the end-effectors) under appropriate force perturbations, and the learned policy must
\emph{encode} compliance up to a reasonable force tolerance limit rather than override it aggressively to maintain good tracking. Consequently, residual schemes
that only edit motor commands or tracked motions~\cite{he2025asap,zhao2025resmimic}
lose the analytical controller's physical bounds. On the other hand, some whole-body trackers and
teleoperation pipelines~\cite{ze2025twist2,li2025clone,lu2025mobiletelevision,fu2024humanplus, ze2025twist}
expose no impedance interface during training, so compliance must be injected separately while running the full analytical pipeline at
deployment. This leads to significantly higher runtime cost and contact-mode fragility on heavy platforms.

We address both challenges with \textsc{CompliantWBC} (\Cref{fig:teaser}).
For the first, we derive an analytical multi-site reference controller
(\Cref{sec:heuristic}) composing centroidal-momentum balance with per-link
Cartesian impedance at arbitrary sites. Rather than cloning its torque, we use
its per-link virtual targets to supervise a base RL policy through a
compliance-fidelity reward~\cite{margolis2025softmimic,
he2025cotapcomplianttaskpipeline}, so the deployed policy can avoid the QP's
runtime cost and contact-mode fragility. For the second, a Phong-weighted
force-origin sampler (\Cref{sec:sampling}) drives perturbations across the whole
body---including pelvis and lower-body sites that upper-body-only curricula never
excite---so the policy \emph{encodes} compliance rather than overriding it. To better introduce external force signal to the policy, we jointly train a
variational \emph{force encoder} with the base policy behind a gradient
barrier from PPO and shaped by wrench reconstruction, contrastive
clustering~\cite{khosla2020supcon}, a KL bottleneck, and temporal smoothness. This \emph{force encoder}
supplies a class-clustered, wrench-grounded estimate of the unobserved contact
state~\cite{demont2024kineticsobservertightlycoupled} and serves as a coarse yet structured signal for the policy. Finally, based on this signal, a bounded
residual on the frozen base edits the impedance \emph{equilibrium} rather than
motor commands or gains~\cite{johannink2019residual, 5991208}.
We study these behaviors on a heavy humanoid platform (\textit{70\,kg}, versus
the \textit{35\,kg} Unitree G1~\cite{unitree_g1} used in most prior
learning-based compliant humanoid work) to
tackle a wider range of contact-rich tasks.
Our contributions can be summarized as follows:
\begin{itemize}
    \item An \textbf{analytical multi-site whole-body compliance reference controller} (\Cref{sec:heuristic}) composing centroidal-momentum balance with per-link Cartesian impedance at arbitrary sites, whose virtual targets guide the compliance-fidelity reward of the base policy.
    \item  A \textbf{bounded residual on the per-link impedance equilibrium}
(\Cref{sec:architecture}) over a frozen base, which corrects the direction- and magnitude-structured error of a
proprioceptive wrench estimate, recovering most of the compliance fidelity of an
oracle wrench.
    \item A \textbf{Phong-weighted force-origin sampler with an
axis-decoupled pelvis anchor} (\Cref{sec:sampling}) that
induces multi-contact, lower-body-inclusive compliance to our policy, evaluated extensively in simulation and demonstrated on a real heavy humanoid (\Cref{sec:exp}).
\end{itemize}
\section{Related Work}
\label{sec:related}
\begin{table}[t]
  \centering
  \vspace*{4pt}%
  \caption{Humanoid whole-body-control landscape. \emph{LB}: lower-body
    compliance (\checkmark), quadruped only (quad.), or none (--).
    \emph{Sites}: where compliance is realized---end-effector (EE), upper body,
    CoM reference, or arbitrary links (whole). \emph{Residual}: where a learned
    correction is applied.}
  \label{tab:landscape}
  \scriptsize
  \setlength{\tabcolsep}{2.6pt}
  \renewcommand{\arraystretch}{0.95}
  \begin{tabular}{@{}lcccc@{}}
    \toprule
    Method                                     & LB         & Sites      & Residual       & Platform     \\
    \midrule
    FALCON~\cite{zhang2025falcon}              & --         & EE         & --             & G1/Booster   \\
    GentleHumanoid~\cite{lu2025gentlehumanoid} & --         & upper      & --             & G1           \\
    FACET~\cite{xu2025facet}                   & quad.      & CoM        & impedance ref. & Go2/G1       \\
    SoftMimic~\cite{margolis2025softmimic}     & per-motion & whole      & --             & G1           \\
    CHIP~\cite{chen2025chip}                   & --         & EE         & hindsight goal & G1           \\
    ResMimic~\cite{zhao2025resmimic}           & --         & --         & on motion      & G1           \\
    ASAP~\cite{he2025asap}                     & --         & --         & on action      & G1           \\
    \midrule
    \textbf{CompliantWBC (ours)}               & \checkmark & \checkmark & on equilibrium & in-house, 70\,kg \\
    \bottomrule
  \end{tabular}
\end{table}

\subsection{Compliance Control for Humanoids}
\label{sec:rel-compliance}

Classical compliance strategies, including task-space impedance and admittance
control~\cite{hogan1985impedance,ott2008cartesian,khatib1987unified,
sentis2005synthesis,henze2016ijrr,dietrich2016wbic,albuschaffer2007cartesian}, regulate
interaction forces via virtual mass--spring--damper dynamics, with
operational-space extensions that handle multiple frames through
hierarchical task-priority projection and a balancing QP that
distributes the centroidal wrench to support contacts. These pipelines
provide rigorous stability guarantees but require accurate dynamics,
hand-crafted task hierarchies, and explicit contact-mode reasoning that
is brittle on heavy humanoid platforms; time-varying stiffness
additionally breaks passivity unless an explicit storage mechanism is
enforced~\cite{ferraguti2013energy,kronander2016stability}.
Recent whole-body tracking controllers~\cite{he2024hover,he2024omnih2o,luo2023phc,kuang2025skillblender} may achieve agile
loco-manipulation via motion imitation but do not expose an impedance
interface, so compliance must be injected separately.

To relax these requirements, recent learning-based methods train
policies to exhibit compliant behavior through reward shaping,
perturbation curricula, per-link virtual springs, or end-effector
stiffness modulation~\cite{zhang2025falcon,lu2025gentlehumanoid,
xu2025facet,chen2025chip,do2025watch}. Most restrict compliance to the end-effector
or the upper-body chain, leaving lower-body bracing, hip translation, and
double-support redistribution---the regimes that dominate heavy
loco-manipulation---unaddressed. Some~\cite{margolis2025softmimic} manage to extend the compliant behavior to the lower body, but limits to a single motion and requires complicated per-motion augmentation to implicitly introduce desired interaction rather than targeting an arbitrary robot configuration within its workspace. A parallel line targets
heavy-payload loco-manipulation with multi-policy or trajectory-optimized
references~\cite{kim2025kinematicsaware,xu2026interactionaware} but similarly does not
expose a per-link impedance interface. Our work composes centroidal-momentum balance
and per-link Cartesian impedance at arbitrary sites inside an RL loop,
admitting end-effector-only and upper-body-only formulations as special cases
(\Cref{tab:landscape}).

\subsection{Residual Policies and Latent Force Estimation}
\label{sec:rel-residual}

Residual reinforcement learning~\cite{johannink2019residual,
silver2018residual,zhang2023residual} composes a learned correction
on top of a hand-crafted base controller, retaining analytical
stability while absorbing unmodeled dynamics. The idea of learning
residuals on controller parameters with RL was pioneered
by Buchli et al.~\cite{buchli2011learning}, who optimized both reference trajectories
and gain schedules on hardware. Our work operates in the same spirit but
residualizes on the impedance \emph{equilibrium} rather than on
gains. Recent humanoid
pipelines instantiate this residual idea differently, focusing on motor
commands, on tracked motions, or on goals via hindsight
relabeling~\cite{he2025asap,zhao2025resmimic,chen2025chip}. However, none of these approaches edit a
per-link impedance interface with an explicit force bound. Consequentially, the residual's
effect on induced contact force is implicit rather than physically bounded. In contrast, we compose per-link
impedance at arbitrary sites and place a \emph{bounded} residual on the
per-link \emph{equilibrium} over a \emph{frozen} base.

A complementary line learns compact latent embeddings of unobserved
context from proprioceptive history, with supervised or contrastive
objectives shown to prevent collapse to motion-phase
memorization~\cite{kumar2021rma,kumar2022arma,lee2020quadrupedal,
oord2018representation,khosla2020supcon}. Recent work specializes such latents
to estimate external wrenches from proprioception without dedicated force
sensors~\cite{lim2023proprioceptive,zhi2025learning,shi2026minimalistcompliance}. We adopt this
proprioceptive estimation approach rather than improve on it: the estimate it supplies
is deliberately coarse, and our contribution is the \emph{bounded residual on the
per-link impedance equilibrium} that it guides. Because the link stiffness $\stiff_\ell$ is held fixed, the residual's effect is
upper-bounded by an explicit virtual-target shift times the stiffness.
\section{Whole-Body Impedance Heuristic}
\label{sec:heuristic}

A humanoid has configuration $\config\in\SE(3)\times\R^{n_j}$, generalized
velocity $\vel$, and floating-base dynamics $\mass\dot\vel + \bvec{C}\vel +
  \bvec{g} = \bvec{S}\trans\torque + \sum_{c}\jac_c\trans\bvec{\lambda}_c +
  \sum_i \jac_{\ell_i}\trans \force_{\text{ext},i}$ with support-contact reaction
wrenches $\bvec{\lambda}_c$ and external wrenches $\force_{\text{ext},i}$ at
links $\ell_i$. Classical Cartesian impedance at a single
end-effector~\cite{hogan1985impedance} prescribes $\mass_d\ddot\pose +
  \damp_d\dot\pose + \stiff_d(\pose-\pose\dsr) = \force_{\text{ext}}$, with
$\pose\in\R^3$. We extend this to a hierarchy of frames coupled by balance.

\textbf{Force-aware virtual targets.}
The single signal carried into training is the per-link virtual target. For a
perturbed link $\ell\in\mathcal{L}$ under an external wrench
$\force_{\text{ext}}$ applied at body site $p$,
\begin{equation}
  \tgt{\ell}{\force_\ell} = \pose_\ell^{\text{ref}} + \stiff_\ell\inv\,\force_\ell,
  \qquad \force_\ell = \mathrm{Ad}_{p\to\ell}^{-\top}\,\force_{\text{ext}},
  \label{eq:vtarget}
\end{equation}
where $\force_\ell$ is the external wrench transported to link $\ell$'s frame and $\stiff_\ell\inv$
is the anisotropic Cartesian compliance, so the correction has units of
displacement. We use the argument throughout to mark a \emph{commanded} target
and to name the wrench it was built from: $\tgt{\ell}{\force}$ uses the
ground-truth wrench and is the reward anchor, $\tgt{\ell}{\hat\force}$ uses the
estimate $\hat\force_{\text{ext}}=\psi(\latent_F)$ and is what the policy is
commanded with, and $\tgt{\ell}{\bvec{0}}=\pose_\ell^{\text{ref}}$ is the nominal
reference. Written without an argument, $\pose_\ell$ is always the
\emph{realized} link position. This generalizes GentleHumanoid's scalar spring to arbitrary
sites and anisotropic stiffness: restricting $\mathcal{L}$ to upper-body links
with $\stiff_\ell\!=\!K_p\bvec{I}_3$, $K_p\!\sim\!\mathcal{U}(5,250)$~N/m and
$\damp_\ell\!=\!2\sqrt{m_v K_p}\bvec{I}_3$ ($m_v\!=\!0.1$~kg) recovers
GentleHumanoid's upper-body Cartesian-spring response~\cite{lu2025gentlehumanoid}
(up to its motion-data-driven guiding contacts), and restricting the perturbed
set to the end-effectors reduces to FALCON's end-effector perturbation
model~\cite{zhang2025falcon} as an analytical special case.
\section{Method: Force-Aware Base Policy and Residual on Impedance Targets}
\label{sec:method}

\begin{figure*}[t]
  \centering
  \includegraphics[width=0.8\textwidth]{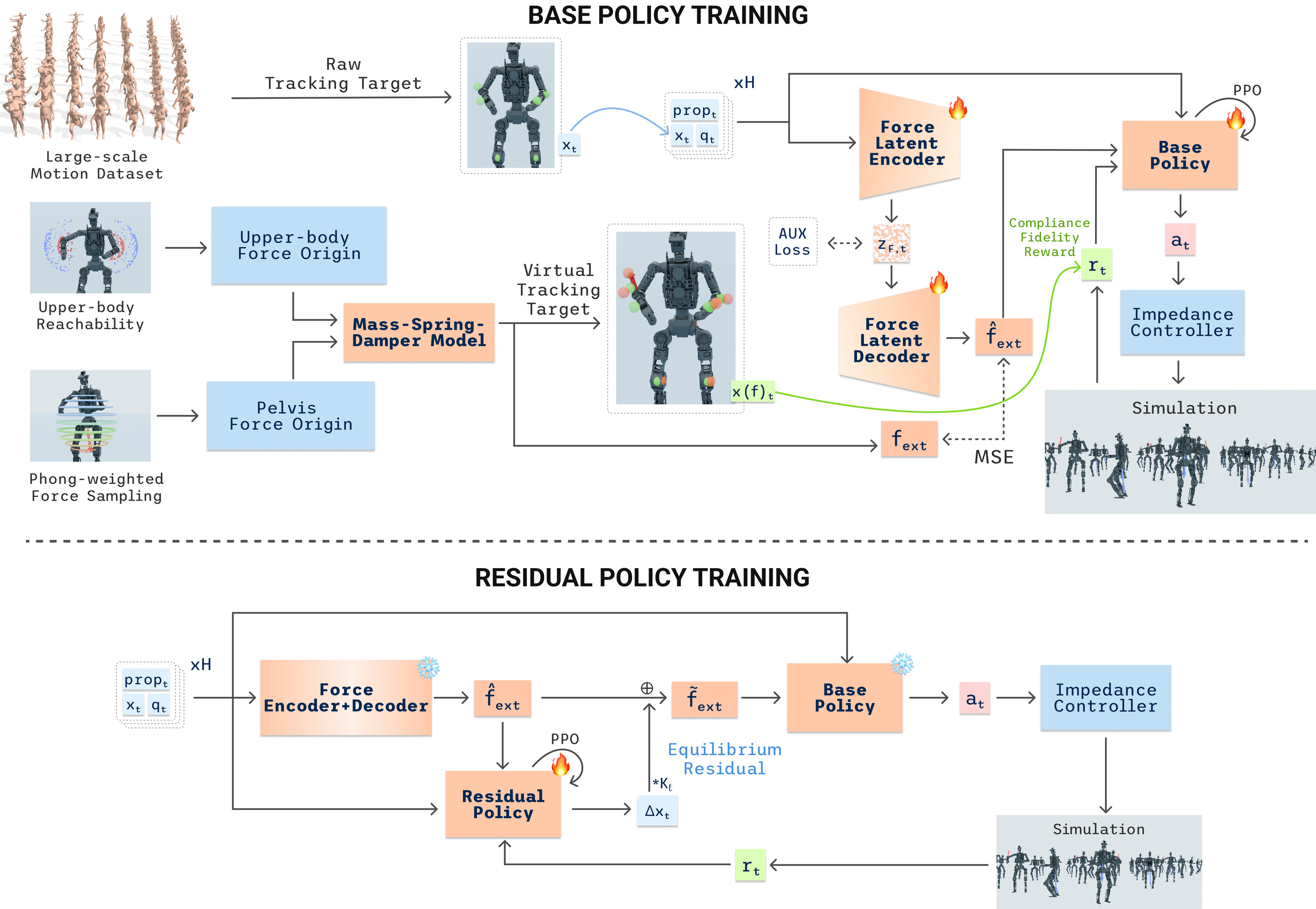}
  \caption{Two-stage training of \textsc{CompliantWBC}: (i) a base policy jointly trained with the force encoder to induce whole-body compliance; (ii) a bounded residual on the impedance equilibrium that compensates for wrench-estimation error.}
  \label{fig:pipeline}
\end{figure*}

\begin{figure}[b]
  \centering
  \includegraphics[width=0.75\columnwidth]{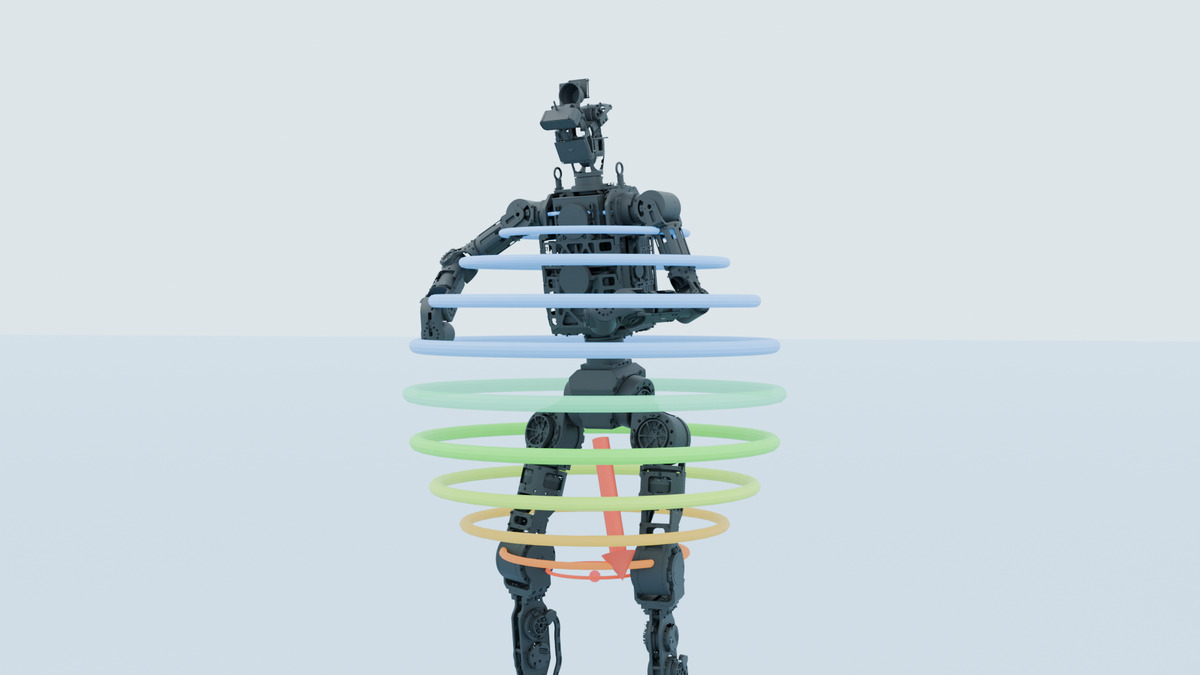}
  \caption{Phong-weighted sampling generates downward-biased force origins,
    decoupled by axis.}
  \label{fig:sampling}
\end{figure}

\subsection{Pipeline Architecture}
\label{sec:architecture}

The analytical controller of \Cref{sec:heuristic} assumes rigid contacts,
perfect torque control, accurate dynamics, and a measured external wrench, none of which a deployed policy on a heavy humanoid can rely on in the rewal world. We therefore
train a two-stage pipeline (\Cref{fig:pipeline}): (i)~a \emph{force-aware base policy}
that jointly trains with a force encoder to induce compliance
behavior, and
(ii)~a \emph{residual policy} on top of the frozen base that compensates for
the discrepancy between the base's wrench estimation and the true
external wrench.

\textbf{Base policy} is a force-aware whole-body policy
$\pi_{\text{base}}(\prop,\config_d,\{\tgt{\ell}{\bvec{0}}\},\hat\force_{\text{ext}})\!\to\!\config^{\rm des}$
that takes as input the proprioception $\prop$, the retargeted
joint targets $\config_d$, the \emph{nominal} per-link targets
$\{\tgt{\ell}{\bvec{0}}\}_{\ell\in\mathcal{L}}$, and the wrench estimate
$\hat\force_{\text{ext}}$ decoded from the force encoder, and outputs
\emph{joint position targets}
$\config^{\rm des}$. A fixed-gain joint-space PD law
$\torque=\stiff_p(\config^{\rm des}-\config)-\damp_p\dot\config$ is the only
torque law executed online; the analytical controller of \Cref{sec:heuristic}
and its balancing QP never run at deployment. The policy is trained end-to-end with
PPO~\cite{schulman2017ppo} against a compliance-fidelity reward
\begin{equation}
r_t = r_{\text{track}}
\;-\;w_c\,\|\pose_\ell-\tgt{\ell}{\force}\|^2
\;-\;w_e\,\|\torque\|^2,
\label{eq:reward}
\end{equation}
whose term $w_c\|\cdot\|^2$ drives the controlled links toward the analytical
controller's virtual-target deformation $\tgt{\ell}{\force}$
(Eq.~\eqref{eq:vtarget}) under the ground-truth perturbation. This
incentivizes the policy to learn multi-site compliance without inheriting the
balancing QP's runtime cost at deployment.

\textbf{Force encoder} follows a variational encoder architecture that maps an observation history window
(proprioceptions, velocities, last torques, and motion command) to a force latent $\latent_F\!\in\!\R^{d_F}$. The encoder is paired with two
auxiliary heads --- a \emph{wrench decoder} $\psi$ that regresses the
per-site external wrench on all controlled links, and a
\emph{projection head} $g$ on the unit sphere dedicated to the
contrastive loss. The encoder is shaped \emph{only} by the auxiliary
objective in Eq.~\eqref{eq:laux} while $\hat\force_{\text{ext}}=\psi(\latent_F)$ is
detached before it is fed to $\pi_{\rm base}$ and $\pi_{\rm res}$. This forms a \emph{gradient barrier} preventing PPO
gradients from flowing through the encoder, so that the latent cannot be co-opted
away from its wrench-grounded objective toward reward-hacking features.

The auxiliary objective $\mathcal{L}_{\phi}$ combines four terms: a per-site
wrench-reconstruction loss (the main term) grounding $\latent_F$ in the true
contact wrench through $\psi$; a supervised contrastive
loss~\cite{khosla2020supcon} over perturbation-class labels (which limb, which
direction); a KL/VIB bottleneck~\cite{alemi2017dvib} suppressing dimensions
that carry neither wrench- nor class-relevant information; and a
temporal-smoothness term on consecutive posterior means:
\begin{equation}
  \mathcal{L}_{\phi}
  =\lambda_w\mathcal{L}_{\rm wrench}
  +\lambda_s\mathcal{L}_{\rm supcon}
  +\lambda_k\mathcal{L}_{\rm kl}
  +\lambda_m\mathcal{L}_{\rm smooth}.
  \label{eq:laux}
\end{equation}
Together they make $\latent_F$ a wrench-grounded, class-clustered, low-bandwidth
embedding of the privileged contact state, which provides a coarse yet structured signal for both the base policy and the residual policy.

\textbf{Residual policy} is a small MLP added on top of the frozen base policy
and frozen force encoder. It reads
$(\prop_t,\config_d,\{\tgt{\ell}{\bvec{0}}\},\hat\force_{\text{ext}})$, with the frozen
wrench estimate $\hat\force_{\text{ext}}\!=\!\psi(\latent_F)$, and outputs a
bounded delta on the impedance equilibrium. The deployed control graph is then
\begin{align}
  \tgte{\ell} & = \tgt{\ell}{\hat\force} + \Delta\pose_\ell,
                \quad \Delta\pose_\ell=\epsilon_x\tanh(\bvec{u}_t),                          \label{eq:res-compose} \\
  \tilde\force_\ell    & = \hat\force_\ell + \stiff_\ell\Delta\pose_\ell,                      \label{eq:dep-wrench}  \\
  \config^{\rm des}_t  & = \pi_{\rm base}(\prop_t,\config_d,\{\tgt{\ell}{\bvec{0}}\},\tilde\force),
                                                                                             \label{eq:dep-base}    \\
  \torque_t            & = \stiff_p(\config^{\rm des}_t-\config_t)-\damp_p\dot\config_t,     \label{eq:dep-pd}
\end{align}
where $\bvec{u}_t=\pi_{\rm res}(\cdot)$ is
the raw residual output and $\Delta\pose_\ell$ the applied edit with
$\|\Delta\pose_\ell\|\!\le\!\epsilon_x\!=\!5$~cm. Because \Cref{eq:vtarget} is affine in
the wrench, $\tgt{\ell}{\hat\force+\stiff_\ell\Delta\pose_\ell}=\tgte{\ell}$, the
equilibrium edit of \Cref{eq:res-compose} and the wrench correction of
\Cref{eq:dep-wrench} are the same residual in two coordinates and therefore can be defined on
the equilibrium and transmitted on $\hat\force_{\text{ext}}$ as policy input.

When $\hat\force_{\text{ext}}$ is biased or noisy with respect to the true
external wrench, the residual edits the impedance \emph{equilibrium} to absorb
the discrepancy without changing the base policy and without modulating gains. Because $\tilde\force$
occupies the same input slot $\pi_{\rm base}$ was trained on in Stage~1, the edit
also stays on the frozen policy's input manifold. Both stages are rewarded
against $\tgt{\ell}{\force}$, the analytical target under the
\emph{ground-truth} wrench and the reward anchor is never the edited
$\tgte{\ell}$, so $\pi_{\rm res}$ cannot earn reward by moving its own
target. We train the residual in a separate stage rather
than jointly with the base, as under a single PPO loss the residual would absorb
compliance the base could itself learn, which would turn its bound into a constraint on
the joint system rather than serve an interpretable discrepancy compensator as in our framework.

\subsection{Compliant virtual targets via external-force sampling}
\label{sec:sampling}

\textbf{Phong-weighted sampling.}
For perturbed upper-body links (wrists, elbows) we sample force origins from a
kinematic reachability cloud and drag the link toward
them~\cite{lu2025gentlehumanoid}. For the pelvis---which endures much larger
forces over a limited range of motion---we instead draw a force origin from a
continuous \emph{Phong-weighted} distribution inside a ball of radius
$R_a\!=\!0.5$~m centered on the pelvis, with direction $\hat{\bvec{n}}$
following a Phong lobe~\cite{phong1975illumination} concentrated on the
downward axis $-\hat{\bvec{e}}_z$ with exponent $n\!=\!2$:
\begin{equation}
  p(\hat{\bvec{n}}) = (1-\epsilon)\,\frac{n+1}{2\pi}\,
  \max\!\bigl(-\hat{\bvec{n}}\cdot\hat{\bvec{e}}_z,\,0\bigr)^{n}
  + \epsilon\,\frac{1}{4\pi},
  \label{eq:phongpdf}
\end{equation}
with isotropic mixing weight $\epsilon\!=\!0.1$, sampled in closed form by
inverse CDF to give the offset
$\bvec{\delta}=r(\sin\theta\cos\phi,\,\sin\theta\sin\phi,\,-\cos\theta)$. The
downward bias emphasizes gravity on a carried load while lateral and upward
samples cover reaching and lifting. Unlike a fixed cloud the distribution is
gap-free, needs no forward-kinematics precomputation, and exposes its bias
through two interpretable parameters $(n,\epsilon)$ (\Cref{fig:sampling}).

\textbf{Axis-decoupled pelvis anchor and compliant height target.}
Anchoring the offset $\bvec{\delta}$ rigidly to the moving pelvis yields a
self-chasing constant load, whereas a world-fixed anchor lets the force grow
without bound as the robot walks away. We instead decouple the anchor by axis:
the horizontal component co-moves with the floating base (leaving locomotion a
bounded, body-relative load), while the vertical component---the axis we want
the policy to yield along---is pinned to the \emph{commanded} base height
$h^*$, giving the pelvis force-origin anchor $\bvec{a}_{\rm pelvis}$:
\begin{equation}
  \bvec{a}_{\rm pelvis}(t) =
  \underbrace{\pose_{\rm pelvis}^{xy} + \bvec{\delta}^{xy}}_{\text{base-relative (horizontal)}}
  \;+\;
  \underbrace{\bigl(h^* + \delta^{z}\bigr)\,\hat{\bvec{e}}_z}_{\text{gravity-anchored (vertical)}},
  \label{eq:anchor-decoupled}
\end{equation}
so the anchor height $z_a=h^*+\delta^z$ is a per-chunk gravity-aligned datum
that does not sink with the robot. We then evaluate the base-height term of
$r_{\text{track}}$ against a compliance-aware target
$z^*_{\rm virt}(t) = h^* + \alpha_z(z_a(t) - h^*)$, where the gain
$\alpha_z = k/(k + k_{\rm leg})$ is set by a two-spring static balance between
the pelvis-anchor stiffness $k$ and the effective vertical leg stiffness
$k_{\rm leg}$: a stiff anchor pulls the comfortable pose
toward the load, a stiff stance resists it. Penalizing
$(z_{\rm pelvis}-z^*_{\rm virt})^2$ thus interpolates from rigid tracking
($\alpha_z\!=\!0$) to full anchor-following ($\alpha_z\!=\!1$), so compliant
lower-body sinking emerges without an explicit compliance offset.
\section{Experiments}
\label{sec:exp}
\begin{table*}[t]
  \centering
  \vspace*{4pt}%
  \caption{Performance of \textsc{CompliantWBC} versus both \emph{compliant} and \emph{non-compliant}
      baselines over 100 paired rollouts. Arrows indicate the preferred direction;
      bracketed columns are in units of $10^{-2}$. Metrics are defined in \Cref{subsec:settings}.}
  \label{tab:sim-ablation}
  \footnotesize
  \setlength{\tabcolsep}{5pt}
  \renewcommand{\arraystretch}{0.95}
  \begin{tabular}{@{}l cc ccc@{}}
      \toprule
                                                    & \multicolumn{2}{c}{Tracking quality}
                                                    & \multicolumn{3}{c}{WB compliance and robustness}                                  \\
      \cmidrule(lr){2-3}\cmidrule(lr){4-6}
      Controller
                                                    & $E_{\rm cmd}^{\rm free}\!\downarrow$
                                                    & $E_{\rm imp}\!\downarrow$
                                                    & $\rho_{\tau}\!\downarrow$
                                                    & $R_{\rm LB}\!\uparrow$
                                                    & $S\!\uparrow$                                                                     \\
                                                    & \scriptsize$[\times 10^{-2}]$
                                                    & \scriptsize$[\times 10^{-2}]$
                                                    & \scriptsize$[\times 10^{-2}]$
                                                    &                                                                                   &        \\
      \midrule
      \textsc{TWIST2}                         & \boldmath{$2.18{\scriptstyle\,\pm\,0.31}$} & $6.79{\scriptstyle\,\pm\,0.72}$ & $2.74{\scriptstyle\,\pm\,0.38}$ & $0.16{\scriptstyle\,\pm\,0.04}$ & $0.59$ \\
      \textsc{GentleHumanoid}                 & $5.22{\scriptstyle\,\pm\,0.47}$ & $3.92{\scriptstyle\,\pm\,0.35}$ & $2.12{\scriptstyle\,\pm\,0.29}$ & $0.14{\scriptstyle\,\pm\,0.03}$ & $0.91$ \\
      \textsc{Ours} w/o pelvis force sampling & $4.12{\scriptstyle\,\pm\,0.38}$ & $4.22{\scriptstyle\,\pm\,0.21}$ & $1.87{\scriptstyle\,\pm\,0.24}$ & $0.16{\scriptstyle\,\pm\,0.03}$ & $0.93$ \\
      \textsc{Ours} w/o residual policy       & $4.29{\scriptstyle\,\pm\,0.35}$ & $3.81{\scriptstyle\,\pm\,0.29}$ & $1.56{\scriptstyle\,\pm\,0.21}$ & $0.20{\scriptstyle\,\pm\,0.03}$ & $0.94$ \\
      \textsc{Ours} (full)                    & $4.65{\scriptstyle\,\pm\,0.41}$ & \boldmath{$2.58{\scriptstyle\,\pm\,0.24}$} & \boldmath{$0.93{\scriptstyle\,\pm\,0.15}$} & \boldmath{$0.31{\scriptstyle\,\pm\,0.04}$} & \boldmath{$0.98$} \\
      \bottomrule
  \end{tabular}
\end{table*}

\textbf{Evaluation goal.}
We ask three questions: (i)~does \emph{whole-body} compliance from multi-site
(wrist, elbow, torso, pelvis, hip, knee) force sampling improve over aggressive tracking and
end-effector-only compliance, especially for forces away from the hands?
(ii)~how accurate is the learned wrench estimate, and what does the bounded
residual on virtual targets actually correct? and (iii)~how does \textsc{CompliantWBC} behave in the
real world on contact-rich tasks? We address (i)--(ii) in \Cref{subsec:sim} and
(iii) in \Cref{subsec:real}.
\subsection{Settings and Metrics}
\label{subsec:settings}
\begin{figure}[t]
  \centering
  \includegraphics[width=0.8\columnwidth]{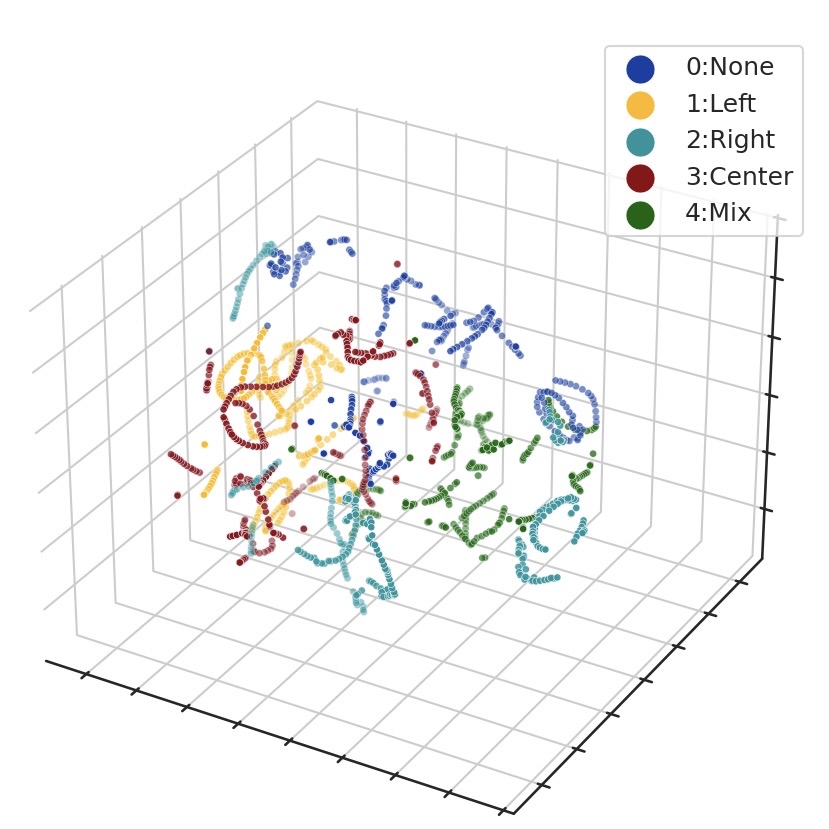}
  \caption{Force latent embedding projections via t-SNE of different force
    profiles across contact sites.}
  \label{fig:latent}
\end{figure}

\begin{figure*}[t]
  \centering
  \includegraphics[width=\textwidth]{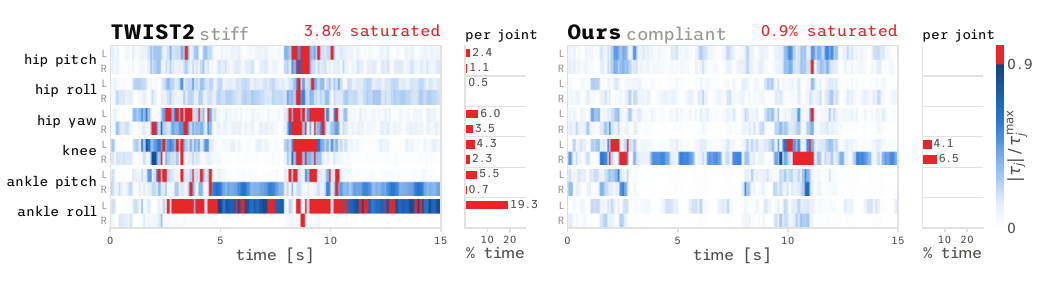}
  \caption{Lower-body joint effort $|\tau_j|/\tau_j^{\max}$ under two trapezoidal pelvis pushes
    (up $1$--$2$\,s and $9$--$10$\,s, down $7$--$8$\,s); L/R rows per joint,
    $0.1$\,s bins. Red: above $0.9\,\tau_j^{\max}$; bars: fraction of time above it.}
  \label{fig:torque}
\end{figure*}

We compare against two external baselines---a stiff whole-body tracker
\textbf{TWIST2}~\cite{ze2025twist2} (nominal command $\pose^{\rm ref}$, no
force-aware targets) and an embodiment-specific re-implementation of
upper-body-only \textbf{GentleHumanoid}~\cite{lu2025gentlehumanoid}---and two
ablations: \textbf{Ours w/o pelvis force sampling} (upper-body perturbations
only, approximating GentleHumanoid-style compliance) and \textbf{Ours w/o
residual} (Stage-1 base policy with the force encoder, no Stage-2
residual). \textbf{Ours (full)} adds the bounded residual on virtual targets with per-link stiffness fixed by the curriculum.
All variants share an identical training recipe, environment budget, and
paired random seeds.

\textbf{Metrics.} Apart from comparing \textbf{Task success $S$} and \textbf{Tracking performance $E_{\rm cmd}^{\rm free}$}, we also report metrics that characterize the compliant response to external forces. \textbf{Torque reaction $\rho_{\tau}$} reflects the fraction of near-saturated joints. A stiff controller fights the force (high $\rho_{\tau}$); a compliant one yields, keeping it low. \textbf{Lower-body participation $R_{\rm LB}$} is the fraction of the total perturbation-induced torque deviation (relative to a matched no-perturbation rollout) borne by the lower-body joints; we treat it as a behavioral \emph{diagnostic} of whether the policy engages the legs, not as a standalone success criterion. We further report \textbf{Compliance fidelity $E_{\rm imp}$}, the distance
between the realized pose of the \emph{perturbed} link and the analytical
impedance response anchored to the \emph{ground-truth} wrench. As the compliant controller deviates from the original tracking target under external forces, $E_{\rm imp}$ compares how close the robot to the analytical controller induced targets so policies that are too stiff or too soft both score badly.
\subsection{Simulation Experiments}
\label{subsec:sim}
\begin{figure*}[t]
  \centering
  \includegraphics[width=\textwidth]{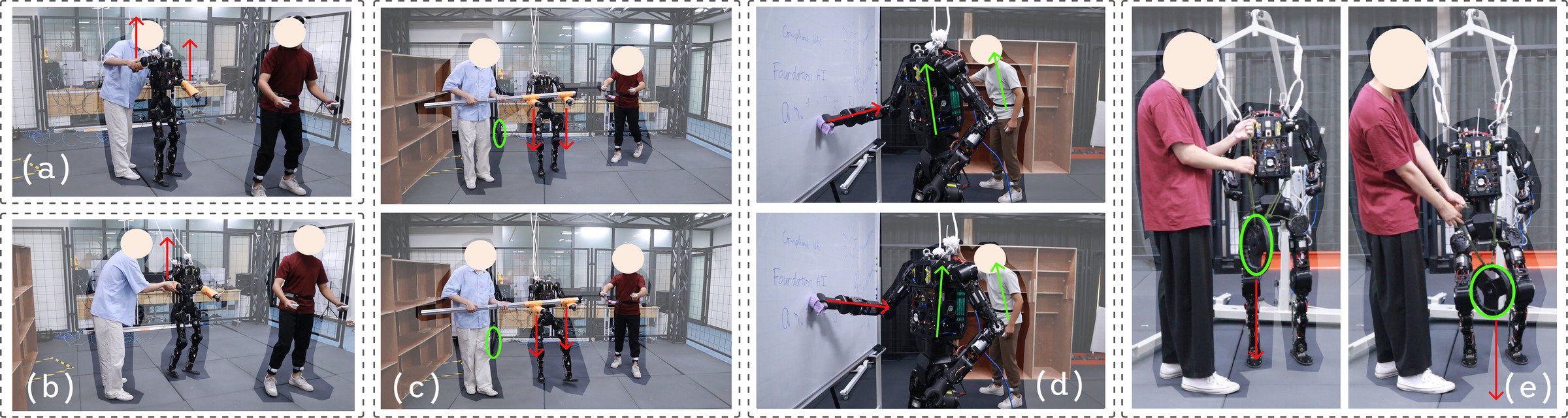}
  \caption{We deploy \textsc{CompliantWBC} in the real world on five force-aware
    tasks: (a) Static Force Reaction; (b) Dynamic Force Reaction; (c) Cooperative
    Payload Transport; (d) Board Wiping; and (e) Squat Under Payload.}
  \label{fig:real}
\end{figure*}

\begin{figure}[b]
  \centering
  \includegraphics[trim=0 3 0 15, clip, width=\columnwidth]{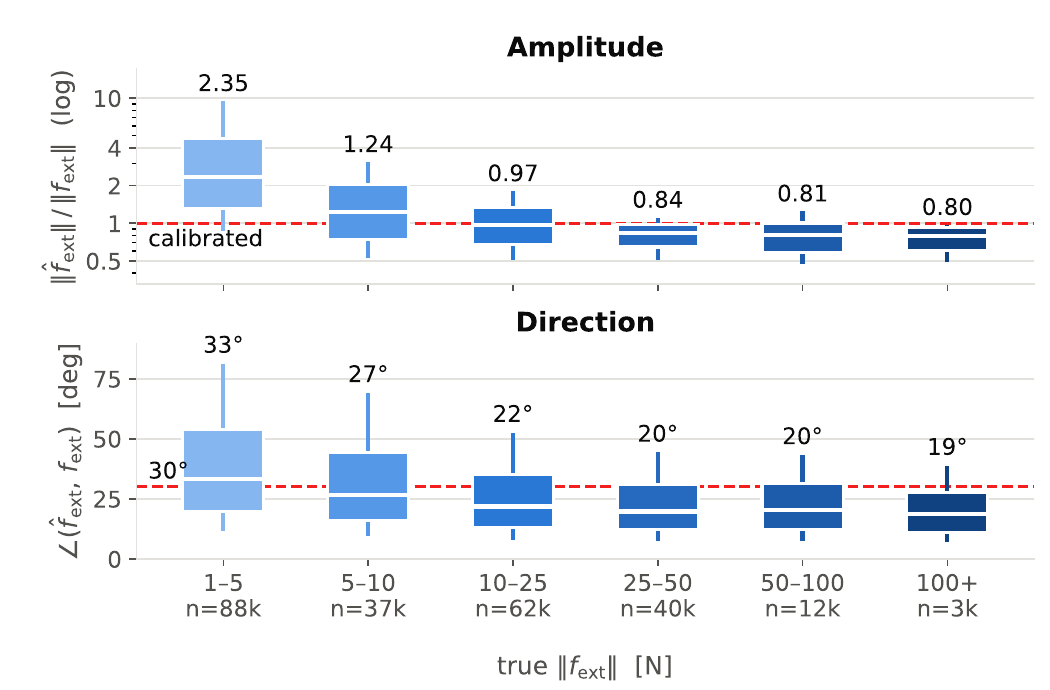}
  \caption{Wrench-estimation quality of the frozen decoder $\psi(\latent_F)$
    against the ground-truth wrench, binned by true load ($n\!=\!242$k
    samples). Direction sharpens monotonically with load, while amplitude
    carries a systematic bias that crosses calibration in the
    $10$--$25$\,N bin.}
  \label{fig:ampdir}
\end{figure}

\begin{table}[t]
  \centering
  \caption{Residual ablation: equilibrium source $\times$ residual
    on/off, over the same 100 paired rollouts. $\hat{\force}$ is the estimated
    wrench, $\force$ the simulator ground truth (oracle); D is \textsc{Ours}
    (full) and A is \textsc{Ours} w/o residual. $\|\Delta\pose\|$ is the mean
    magnitude of the residual's equilibrium edit.}
  \label{tab:residual}
  \footnotesize
  \setlength{\tabcolsep}{3pt}
  \begin{tabular}{@{}l ccccc@{}}
    \toprule
    Variant                   & $\|\force-\hat{\force}\|$      & $E_{\rm imp}\!\downarrow$                  & $\rho_{\tau}\!\downarrow$                  & $\|\Delta\pose\|$ & $S\!\uparrow$ \\
                              & \scriptsize[N]                 & \scriptsize[cm]                            & \scriptsize$[\times 10^{-2}]$              & \scriptsize[cm]   &               \\
    \midrule
    A\; ($\hat{\force}$, off) & $10.1{\scriptstyle\,\pm\,2.1}$ & $3.81{\scriptstyle\,\pm\,0.29}$            & $1.56{\scriptstyle\,\pm\,0.21}$            & --                & $0.94$        \\
    B\; ($\force$, off)       & $0$                            & $2.44{\scriptstyle\,\pm\,0.22}$            & $1.02{\scriptstyle\,\pm\,0.17}$            & --                & $0.97$        \\
    C\; ($\force$, on)        & $0$                            & \boldmath{$2.31{\scriptstyle\,\pm\,0.21}$} & \boldmath{$0.89{\scriptstyle\,\pm\,0.14}$} & $0.4$             & $0.98$        \\
    D\; ($\hat{\force}$, on)  & $10.1{\scriptstyle\,\pm\,2.1}$ & $2.58{\scriptstyle\,\pm\,0.24}$            & $0.93{\scriptstyle\,\pm\,0.15}$            & $2.1$             & $0.98$        \\
    \bottomrule
  \end{tabular}
\end{table}

\textbf{How compliant is the policy?} \Cref{tab:sim-ablation} lays out the
tracking--compliance trade-off. \textsc{TWIST2} tracks best in free space but
succeeds on only $59\%$ of trials, fighting every perturbation. This is also demonstrated by the worst impedance fidelity score $E_{\rm imp}=6.79$.
\textsc{GentleHumanoid} reaches $S\!=\!0.91$ by relaxing the upper-body limbs
but shows the lowest whole-body engagement ($R_{\rm LB}\!=\!0.14$).
\textsc{Ours (full)} sits on the favorable frontier, with the highest
lower-body participation and the lowest joint saturation $\rho_{\tau}$ at the
highest success rate, at only a modest free-tracking gap to \textsc{TWIST2}. It also attains the best compliance fidelity ($E_{\rm imp}\!=\!2.58$, against $\geq\!3.81$ for every baseline and ablation).
The ablations isolate each component: removing pelvis sampling collapses
$R_{\mathrm{LB}}$ toward the \textsc{GentleHumanoid} regime, while removing
the residual recovers slight tracking precision at the
cost of compliance and robustness, demonstrated by worst joint saturation, lower-body participation, success rate.
We highlight the distinct behavior in the lower body between the whole-body compliant and non-compliant policy under a pelvis perturbation in \Cref{fig:torque}. The compliant policy keeps lower-body torques much smaller
and stays stable through the contact, especially when the force is saturated at its peak in the trapezoid, consistent with its lower $\rho_{\tau}$
in \Cref{tab:sim-ablation} and concretizing the resist-and-fail behavior behind
\textsc{TWIST2}'s low success rate.

\textbf{How accurate is the wrench estimate?} \Cref{fig:ampdir} evaluates the
frozen decoder $\hat{\force}_{\rm ext}=\psi(\latent_F)$ against the ground-truth
wrench over $242$k evaluation samples, separated into \emph{direction} and
\emph{amplitude}. From the the figure, direction sharpens monotonically with load: the median
$\angle(\hat{\force},\force)$ falls from $33^\circ$ in the $1$--$5$\,N bin to
$19^\circ$ above $100$\,N. This suggests that larger force induces sharper signal to the policy, guiding the robot to which direction it should yield to. Amplitude instead carries a systematic, monotone
bias --- the median ratio $\|\hat{\force}\|/\|\force\|$ runs
$2.35\!\rightarrow\!0.97\!\rightarrow\!0.80$ across the same range, crossing
calibration in the $10$--$25$\,N bin and under-estimating heavy loads
thereafter. The force estimation analysis along the direction and magnitude axis together characterizes the residual's job: the estimate
is largely \emph{collinear} with the true wrench, so the policy yields in the
right direction but by the wrong distance, leaving an equilibrium displaced
along a \emph{known} axis by an unknown amount. This is exactly the error
structure a bounded edit on the equilibrium can absorb. We provide further qualitative insight into the learned behavior of the force encoder and the compliant response of the policy. \Cref{fig:latent} shows the force encoder organizes its embeddings into distinct per-site clusters (left/right limb, pelvis, mixed, and
no-perturbation)---a force-origin-aware representation rather than a collapsed
force/no-force mode, a prerequisite for site-dependent reactions.

\textbf{What does the residual correct?} \Cref{tab:residual} crosses the
equilibrium source (estimated $\hat{\force}$ vs.\ oracle $\force$) with the
residual (off/on), which ablates the contribution of our residual policy introduced in Stage~2. If the
residual compensates \emph{wrench-estimation error}, it should help only when
the equilibrium is built from $\hat{\force}$; if it is generic adaptation, it
should help equally either way. The data support the estimation error compensation nature of the Residual: enabling it is
worth $1.23$\,cm of $E_{\rm imp}$ and $0.63$ of $\rho_{\tau}$ under the
estimated wrench (A$\,\to\,$D), but only $0.13$ on each under the oracle
(B$\,\to\,$C). Its own output shrinks accordingly --- the mean equilibrium edit
$\|\Delta\pose\|$ falls $5.3\times$, from $2.1$\,cm (D) to $0.4$\,cm (C) ---
so with nothing left to correct the residual largely switches itself off. In
absolute terms it recovers $90\%$ of the oracle gap on $E_{\rm imp}$
(A $3.81\!\to\!$ D $2.58$, oracle B $2.44$) and closes it entirely on
$\rho_{\tau}$ ($1.56\!\to\!0.93$, against B $1.02$). Access to the true wrench
is therefore worth little once the residual is present, which is why the coarse
estimator of \Cref{fig:ampdir} provides sufficient information for the residual policy to react.

\subsection{Real-World Experiments}
\label{subsec:real}
We deploy our trained policy on a real humanoid via
teleoperation~\cite{xrotbot2025, araujo2025retargeting, pico2023} on five
tasks: \textit{Static Force Reaction}, \textit{Dynamic Force Reaction},
\textit{Cooperative Payload Transport}, \textit{Board Wiping}, and
\textit{Squat Under Payload} (\Cref{fig:real}). The first two mirror the
simulation perturbation protocol; the remaining three apply sustained,
structured contacts that no single-site end-effector formulation can absorb,
directly stressing the multi-site, whole-body assumptions of our pipeline
(\Cref{sec:method}).

Under large pushes while tracking a dynamic walking motion (\Cref{fig:real}b,c)
the robot stays balanced---the regime where the stiff \textsc{TWIST2} collapses
in simulation ($S\!=\!0.59$). In \textit{Cooperative Payload Transport} the
operator hands over a $100\,\mathrm{N}$ bar; the robot yields at the wrists,
redistributes the load through torso and hips, and re-establishes a stable
gait---the yield-then-reconcile behavior of the impedance equilibrium
(Eq.~\eqref{eq:vtarget}). \textit{Squat Under Payload} stresses the same
mechanism vertically, spreading sustained pelvis load across the chain rather
than saturating the hip and knee. \textit{Board Wiping} is the clearest
whole-body case: as the operator presses a board against the robot's hand, the
humanoid leans its trunk back and shifts its CoM over the support
polygon---a response an upper-body-only policy is not trained for, since no
perturbation reaches its lower body in training. These demonstrations are
qualitatively consistent with the simulation results.

\section{CONCLUSIONS}

We presented \textsc{CompliantWBC}, a two-stage RL pipeline
that defines a multi-site whole-body impedance reference controller and uses it
as a compliance-fidelity reward signal. Stage~1 jointly trains the force encoder and a force-aware base policy against this compliance-fidelity reward while stage~2 trains a
bounded residual on the impedance equilibrium that compensates for the
discrepancy between the frozen wrench estimator and the true external
wrench. We validate the result on
heavy-payload loco-manipulation tasks that require lower-body and multi-contact
compliance, observing compliance behavior with better whole-body compliance with a significantly higher lower-body participation both in simulation and in the real world.

\bibliographystyle{IEEEtran}
\bibliography{IEEEabrv,reference}

\end{document}